\documentclass{article} 
\usepackage{iclr2027_conference,times}

\usepackage{amsmath,amsfonts,bm}

\def\eqref#1{equation~\ref{#1}}

\def\1{\bm{1}}

\DeclareMathAlphabet{\mathsfit}{\encodingdefault}{\sfdefault}{m}{sl}
\SetMathAlphabet{\mathsfit}{bold}{\encodingdefault}{\sfdefault}{bx}{n}

\usepackage{hyperref}
\usepackage{url}
\usepackage[dvipsnames]{xcolor}
\usepackage{graphicx}
\usepackage{wrapfig}
\usepackage{booktabs}
\usepackage{multirow}
\usepackage{makecell}
\usepackage{tabularx}
\usepackage{array}
\usepackage{adjustbox}

\newcolumntype{Y}{>{\centering\arraybackslash}X}

\title{LighTROcc: Lightweight 4D Occupancy Forecasting via Instance-Centric 3D Gaussians}

\author{
Hwanhee Jung\textsuperscript{1},
SeungHyeon Kim\textsuperscript{1},
Inkyu Koo\textsuperscript{1},
Qixing Huang\textsuperscript{2},
Sang Ho Yoon\textsuperscript{3},
Sangpil Kim\textsuperscript{1}
\\
{\normalfont
\textsuperscript{1}Korea University \qquad
\textsuperscript{2}University of Texas at Austin
}
\\
{\normalfont
\textsuperscript{3}Korea Advanced Institute of Science \& Technology
}
}

\iclrfinalcopy 

\begin{document}

\maketitle

\lhead{} 

\begin{abstract}
Forecasting future 3D occupancy from surround-view cameras is essential for autonomous driving, yet existing approaches rely on dense voxel or bird's-eye-view representations whose cost grows rapidly with spatial resolution and prediction horizon. Because these representations do not explicitly maintain object identities, they also struggle to preserve instance consistency over time. We present \textbf{LighTROcc}, a lightweight instance-centric framework that represents movable objects with a compact set of learned queries and predicts present and future occupancy in a single forward pass. LighTROcc localizes each query through attention-guided forward lifting, combining image-space cross-attention, query-specific depth, and camera geometry to estimate its 3D center. Each instance is modeled as a mixture of anisotropic 3D Gaussians and propagated across future steps using predicted displacements, producing continuous, temporally consistent occupancy forecasts. Experiments on nuScenes and supplemented nuScenes-Occupancy show that LighTROcc outperforms the evaluated dense and instance-wise baselines in instance-level forecasting accuracy while maintaining strong voxel-level occupancy quality. Across different model configurations, LighTROcc achieves a favorable balance between forecasting accuracy and computational efficiency, demonstrating the potential of compact instance-centric modeling for camera-based 4D occupancy forecasting.
\end{abstract}

\vspace{-1.3em}
\begin{figure}[h]
    \centering
    \includegraphics[width=0.9\textwidth]{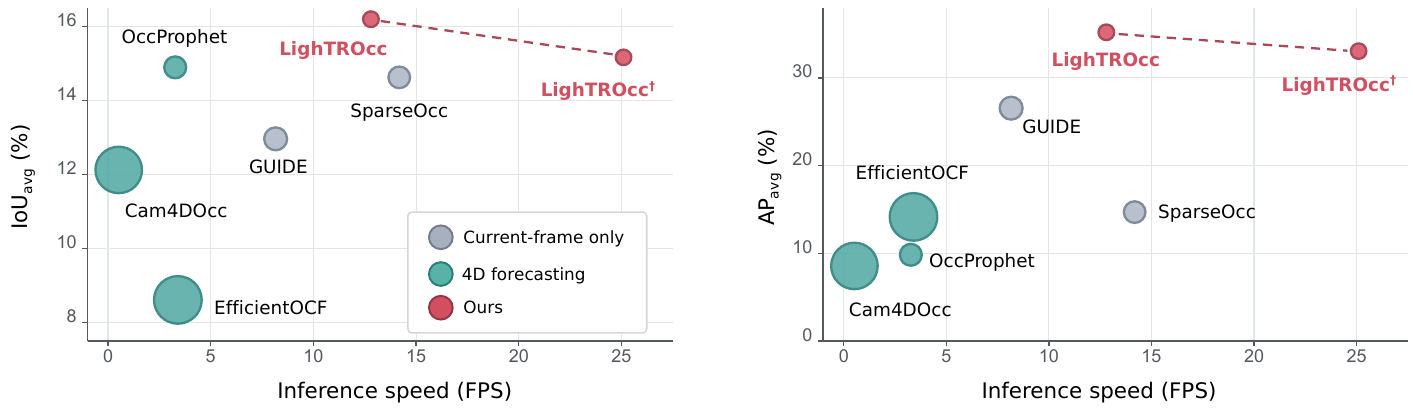}
    \vspace{-1.0em}
    \caption{
        Accuracy-efficiency comparison with current-only and 4D forecasting methods. LighTROcc achieves a favorable trade-off between inference speed and both IoU and AP while predicting current and future occupancy in a single forward pass.
    }
    \label{fig:teaser}
\end{figure}

\section{Introduction}
\label{sec:introduction}

Safe and real-time autonomous driving requires an accurate and efficient understanding of which regions of the surrounding space are free or occupied. Extensive research has investigated fine-grained camera-based occupancy perception to provide a comprehensive representation of the current driving environment~\citep{huang2023tpvformer,wei2023surroundocc,zhang2023occformer,wang2023openoccupancy,Tong_2023_ICCV,ma2024cotr}. From the perspective of planning and control, however, estimating only the present state is insufficient. An autonomous vehicle must also anticipate whether currently free space will remain traversable or become occupied when the vehicle reaches it. This need has motivated growing interest in 4D occupancy forecasting, which predicts a sequence of future 3D occupancy states from past observations. Since temporal occupancy changes are primarily driven by movable objects, these forecasting methods commonly focus on general movable objects (GMOs).

Recent 4D occupancy forecasting methods typically model temporal evolution by extending spatially indexed representations, such as voxels or bird's-eye-view (BEV) cells, along the temporal dimension~\citep{ma2024cam4docc, xu2025efficientocf, chen2025occprophet}. Many of these methods rely on volumetric processing, such as 3D convolutions, to reconstruct and forecast occupancy. Although such representations retain an explicit volumetric description of the scene, they require substantial computation, and their cost increases rapidly with spatial resolution and forecasting horizon~\citep{chen2025occprophet}. Consequently, forecasting multiple future frames while satisfying real-time inference requirements remains challenging. A recent approach factorizes 3D occupancy into a 2D BEV representation and column-wise height information to avoid explicit full-volume processing~\citep{xu2025efficientocf}. While this lower-dimensional formulation can alleviate the burden of volumetric processing, its column-based reconstruction may inadequately capture complex object geometry and fine-grained volumetric extents. More broadly, existing 4D occupancy forecasting methods are generally not designed in an instance-centric manner. When object identities are required, instances must instead be recovered through additional association, grouping, or flow-based processing. Such indirect recovery can lead to temporal inconsistencies, including the merging of distinct objects or the fragmentation of a single object across frames.

Maintaining consistent object identities over time therefore calls for an instance-centric formulation that directly recognizes and forecasts individual objects. Existing approaches have explored instance query candidates, where each query is intended to represent an object~\citep{wang2022detr3d, liu2022petr, liu2024sparseocc, hu2026guide}. In many such methods, queries are initialized from spatial anchor points, and a large candidate set is required to provide sufficient coverage of the 3D scene. Because this set is predefined to be substantially larger than the actual number of object instances, it can produce redundant predictions and incur unnecessary computational overhead. Furthermore, these queries are generally designed to represent instances in the current scene rather than to persist as object-specific states throughout future frames. These observations motivate a lightweight instance-centric framework that compactly represents the current scene and directly propagates object instances across multiple future time steps.

Building on this principle, we introduce \textbf{LighTROcc}, a lightweight instance-centric transformer for camera-based 4D occupancy forecasting. Rather than distributing a large set of spatially initialized candidate queries throughout the 3D scene, LighTROcc directly localizes a compact set of queries in 3D through attention-guided forward lifting. Specifically, each query derives its image-space location from the attention map and combines it with a query-specific depth prediction and camera geometry. The localized query further predicts a mixture of 3D Gaussians that compactly represents the full volumetric extent of the corresponding instance. The resulting instance representation is then propagated to future frames using predicted displacements, maintaining object-level consistency over time. This compact formulation enables LighTROcc to forecast multiple future occupancy states in a single forward pass, yielding a favorable balance between forecasting accuracy and inference speed, as shown in Figure~\ref{fig:teaser}. We evaluate our method on nuScenes~\citep{caesar2020nuscenes} and supplemented annotations from nuScenes-Occupancy~\citep{wang2023openoccupancy}, where it outperforms the evaluated baselines, demonstrating the effectiveness of the proposed formulation.

Our contributions are summarized as follows:
\begin{itemize}
    \item We propose LighTROcc, a lightweight instance-centric network that forecasts future 4D occupancy states in a single forward pass.
    \item We introduce an attention-guided forward lifting formulation that directly localizes 3D instance centers from attention map and depth, enabling compact instance prediction.
    \item Extensive experiments on nuScenes and supplemented nuScenes-Occupancy demonstrate the highest accuracy with lower computational cost and faster inference.
\end{itemize}


\section{Related Work}
\label{sec:related_work}

\subsection{3D Occupancy Representations and Query-Based Perception}
\label{sec:related_queries}

\vspace{-0.6em}
\paragraph{Occupancy representations.}
3D occupancy prediction has been explored using dense voxel grids, reduced-dimensional representations, and continuous 3D primitives. Early semantic scene completion and camera-based methods construct dense voxel volumes to preserve full 3D structure, but their computational cost scales with spatial resolution~\citep{song2017sscnet,roldao2020lmscnet,cao2022monoscene,wei2023surroundocc,li2023fbocc}. To reduce this burden, FlashOcc compresses volumetric occupancy by mapping BEV features to height-aware outputs, while PointOcc factorizes 3D space into cylindrical tri-perspective representations~\citep{yu2023flashocc,zuo2023pointocc}. More recently, Gaussian-based approaches provide continuous alternatives to voxel grids using sparse 3D Gaussians or Gaussian splatting representations for occupancy prediction~\citep{huang2024gaussianformer,huang2025gaussianformer2,gan2025gaussianocc,jiang2025gausstr}.

\vspace{-0.6em}
\paragraph{Geometry-indexed query methods.}
A major query-based line assigns queries to predefined geometric elements such as BEV cells, planes, or voxels. BEVFormer learns dense BEV queries, TPVFormer introduces three orthogonal query planes, and VoxFormer and PanoOcc adopt voxel-indexed representations for occupancy prediction~\citep{li2022bevformer,huang2023tpvformer,li2023voxformer,wang2024panoocc}. Subsequent methods improve efficiency through perspective supervision, polar layouts, dual-path decoding, octree queries, and compact query representations~\citep{yang2023bevformerv2,jiang2023polarformer,zhang2023occformer,lu2024octreeocc,ma2024cotr,oh2025protoocc}. However, spatially indexed queries remain coupled to the predefined discretization, causing computational and memory costs to grow with spatial resolution.

\vspace{-0.6em}
\paragraph{Object-centric query methods.}
DETR3D associates queries with 3D reference points projected into camera views, while PETR and PETRv2 incorporate 3D positional information into image features~\citep{wang2022detr3d,liu2022petr,liu2023petrv2}. Subsequent methods extend sparse queries to temporal modeling, tracking, sensor fusion, and end-to-end driving~\citep{lin2022sparse4d,lin2023sparse4dv2,lin2023sparse4dv3,liu2023sparsebev,wang2023streampetr,xiong2023cape,zhang2022mutr3d,chen2023futr3d,sun2025sparsedrive}. Nevertheless, several approaches still rely on reference points or anchor-based initialization, requiring large candidate sets for scene coverage~\citep{wang2022detr3d,lin2022sparse4d}. Sparse query formulations have also been extended to occupancy prediction through sparse points, sets, contextual queries, and Gaussian representations~\citep{tang2024sparseocc,liu2024sparseocc,wang2024opus,shi2024osp,jiang2024symphonize,hu2026guide,Park2026S2GO}. S2GO uses spatially anchored 3D queries with Gaussian decoding for streaming occupancy, whereas LighTROcc adopts compact object-centric queries with attention-guided localization for multi-step occupancy forecasting.

\subsection{Camera-Only 4D Occupancy Forecasting}
\label{sec:related_forecasting}
\vspace{-0.4em}
Camera-only 4D occupancy forecasting predicts present and future 3D occupancy states from historical multi-view images. Cam4DOcc introduced a benchmark and OCFNet, an end-to-end baseline that processes multi-frame 3D voxel features~\citep{ma2024cam4docc}. OccProphet improves efficiency with a lightweight Observer-Forecaster-Refiner framework that aggregates multi-frame voxel features~\citep{chen2025occprophet}, while EfficientOCF decouples spatial and temporal processing through BEV occupancy, height, and flow prediction~\citep{xu2025efficientocf}. Despite these advances, their forecasts remain spatially indexed rather than maintaining objects as persistent instance states. LighTROcc instead represents each object with an instance-specific query and forecasts its occupancy across future frames, preserving object-level consistency throughout the forecasting horizon.

\vspace{-0.5em}
\section{Method}
\label{sec:method}

\vspace{-0.4em}
\subsection{Problem formulation}
\label{sec:problem_formulation}

\vspace{-0.4em}
Let $\bm{I}_t^v$ denote the RGB image captured by camera $v \in \{1,\ldots,V\}$ at time $t \in \{-T_p,\ldots,0\}$, where $T_p$ is the number of past frames preceding the present and $t=0$ denotes the present frame. Following prior camera-only occupancy forecasting methods~\citep{ma2024cam4docc,xu2025efficientocf,chen2025occprophet}, we focus on the occupancy evolution of general movable objects (GMOs). Given the multi-view image history and the corresponding camera intrinsics, extrinsics, and ego poses, the task is to predict the present and future voxel occupancy states:
\begin{equation}
    \{\hat{\bm{O}}_k\}_{k=0}^{T_f},
    \qquad
    \hat{\bm{O}}_k \in [0,1]^{X \times Y \times Z}.
\end{equation}
Here, $\hat{\bm{O}}_k(x,y,z)$ denotes the probability that voxel $(x,y,z)$ is occupied at time $k$, where $k=0$ corresponds to the present and $k>0$ to the future. We set $T_p=2$ and forecast $T_f=4$ frames.

Unlike conventional dense approaches that construct voxel-wise features over a discretized 3D grid, LighTROcc represents movable objects with a compact set of instance queries. Each query predicts confidence, depth, Gaussian mixture parameters, and a future trajectory. The resulting continuous Gaussian volumes are rasterized into the occupancy sequence $\{\hat{\bm{O}}_k\}_{k=0}^{T_f}$.

\subsection{Transformer-based Query Decoder}
\label{sec:transformer_decoder}

\begin{figure}[t]
    \centering
    \includegraphics[width=\textwidth]{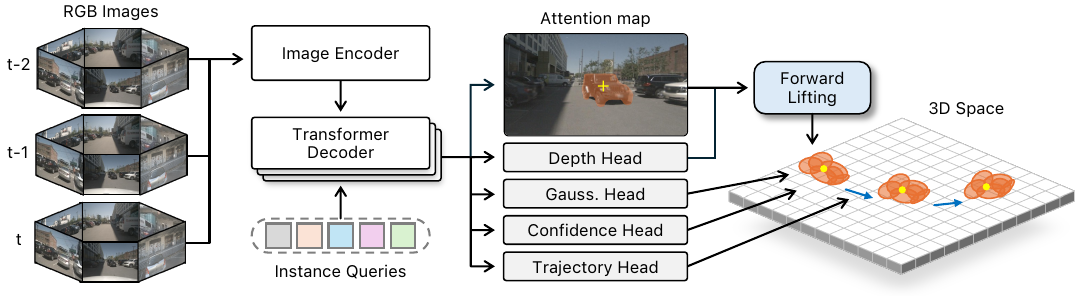}
    \caption{
        Overview of LighTROcc. Compact instance queries are sequentially refined through a transformer decoder using multi-view image features. Each query is localized in 3D via attention-guided forward lifting, while heads predict 3D Gaussians, confidence, and future trajectory.
    }
    \vspace{-1.2em}
    \label{fig:architecture}
\end{figure}

As illustrated in Fig.~\ref{fig:architecture}, a shared image encoder extracts multi-scale context features $\bm{F}_{t, \ell}^v \in \mathbb{R}^{H_{\ell} \times W_{\ell} \times D}$ from each surrounding image $\bm{I}_t^v$. Since the shared encoder processes each image independently, the resulting visual features do not explicitly encode their spatial locations, camera viewpoints, or acquisition times. We therefore augment each image token with a 2D sinusoidal positional encoding, a camera embedding derived from the corresponding camera parameters and ego pose, a learned camera-ID embedding, and a temporal embedding representing the observation time. For decoder layer $\ell$, the augmented tokens from all cameras are flattened to form the multi-view context features $\bm{F}_{t,\ell}$.

We initialize the decoder with $N_q$ learned instance queries $\bm{Q} \in \mathbb{R}^{N_q \times D}$. For each observed frame, the queries sequentially interact with the multi-view features through cross-attention and are refined by self-attention and a feed-forward network. At decoder layer $\ell$, the update is formulated as

\vspace{-1.em}
\begin{equation}
\begin{aligned}
\bar{\bm{Q}}_{t}^{\ell} & = \operatorname{LN}\!\left(\bm{Q}_t^{\ell-1}+ \operatorname{CA}_{\ell}(\bm{Q}_t^{\ell-1},\bm{F}_{t,\ell})\right),\\ \widetilde{\bm{Q}}_{t}^{\ell} & = \operatorname{LN}\!\left(\bar{\bm{Q}}_{t}^{\ell}+ \operatorname{SA}_{\ell}(\bar{\bm{Q}}_{t}^{\ell})\right),\\ \bm{Q}_t^{\ell} & = \operatorname{LN}\!\left( \widetilde{\bm{Q}}_{t}^{\ell}+ \operatorname{FFN}_{\ell}(\widetilde{\bm{Q}}_{t}^{\ell})\right).
\end{aligned}
\end{equation}
\vspace{-1.em}

Here, $\operatorname{LN}$ denotes layer normalization, while $\operatorname{CA}_{\ell}$, $\operatorname{SA}_{\ell}$, and $\operatorname{FFN}_{\ell}$ denote the cross-attention, self-attention, and feed-forward modules at layer $\ell$, respectively.
The observation frames are processed chronologically, with the final query state of each frame initializing the next. This sequential propagation serves as a compact temporal memory without constructing dense spatiotemporal features.

For efficient coarse-to-fine decoding, successive layers attend to image features of increasing spatial resolution. Early layers capture broad context at low computational cost, while the final layer retains the spatial detail required for 3D lifting. We use three decoder layers with four attention heads.

\begin{wrapfigure}[12]{r}{0.50\textwidth}
    \centering
    \vspace{-0.5em}
    \includegraphics[width=\linewidth]{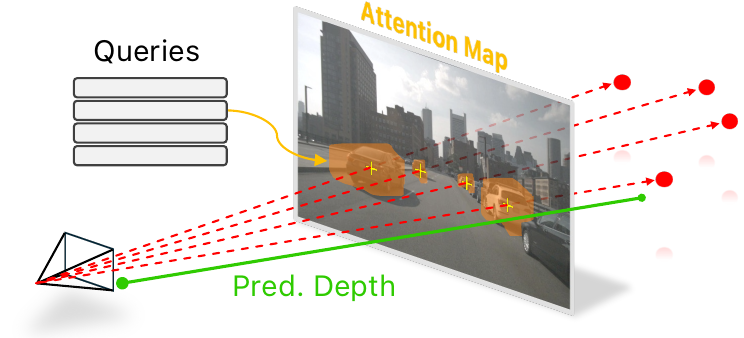}
    \vspace{-1.7em}
    \caption{Illustration of attention-guided forward lifting, combining query attention and predicted depth to recover each instance center in 3D.}
    \label{fig:lifting}
\end{wrapfigure}

\subsection{Attention-guided forward lifting}
\label{sec:forward_lifting}

After query decoding, each query contains object-specific 2D visual evidence, but directly regressing its metric 3D location from the latent query vector remains non-trivial. We therefore decompose 3D localization into two simpler predictions: image-plane localization and query-specific depth estimation. The known camera parameters and ego pose then serve as a geometric bridge that lifts these predictions into 3D. Specifically, as illustrated in Figure~\ref{fig:lifting}, we reuse the query-to-image attention as an image-plane localization cue and combine it with the predicted depth to recover a 3D center.

Let $\bm{A}_{t,q}^{v} \in [0,1]^{H \times W}$ denote the final-layer cross-attention map of query $q$ over camera $v$ at time $t$. We spatially normalize the attention map within each view and obtain a differentiable image-plane center through soft-argmax:

\vspace{-1.em}
\begin{equation}
\bm{p}_{t,q}^{v}
    =
    \sum_{\bm{u} \in \Omega}
    \bar{\bm{A}}_{t,q}^{v}(\bm{u})\,\bm{u},
\end{equation}
\vspace{-0.8em}

where $\Omega$ denotes the spatial domain of the feature map, $\bm{u}=[u,v]^\top$ is a feature-map coordinate, and $\bar{\bm{A}}_{t,q}^{v}$ is the normalized attention map. To preserve gradients and stabilize image-plane localization, we use soft-argmax rather than discrete argmax. In parallel, a lightweight depth head predicts a categorical distribution $\bm{\pi}_{t,q} \in [0,1]^{N_d}$ over $N_d$ predefined depth bins. Given a depth range $[d_{\min},d_{\max}]$, the center of the $b$-th bin is defined as

\vspace{-0.8em}
\begin{equation}
\rho_b=d_{\min}+\left(b-\frac{1}{2}\right)\frac{d_{\max}-d_{\min}}{N_d},\qquad b=1,\ldots,N_d,
\end{equation}
\vspace{-0.8em}

and the continuous query depth is obtained as the expectation over the fixed bin centers:

\vspace{-1.em}
\begin{equation}
d_{t,q}=\sum_{b=1}^{N_d}\pi_{t,q,b}\rho_b.
\end{equation}
\vspace{-0.8em}

The attention map determines the image region associated with the query, while the depth distribution resolves its position along the corresponding camera ray. For each query, we select the valid camera with the largest total mass in the original cross-attention map and map its location back to the original image coordinates. The resulting image-plane location is then lifted into 3D using the predicted depth and the corresponding camera geometry:

\vspace{-1.em}
\begin{equation}
\bm{c}_{t,q} = \operatorname{Lift}
\left(\bm{p}_{t,q}, d_{t,q}; \bm{K}_{t}, \bm{E}_{t} \right),
\end{equation}

where $\bm{K}_{t}$ and $\bm{E}_{t}$ denote the intrinsic and extrinsic parameters of the selected camera, respectively. The $\operatorname{Lift}(\cdot)$ operation maps the image-plane location to 3D along its camera ray using the predicted depth and camera geometry. This attention-guided forward lifting provides each query with a consistent 3D center without introducing spatially initialized 3D candidates.

\subsection{Occupancy Forecasting}
\label{sec:occupancy_forecasting}

To capture finer object geometry than the coarse occupancy volumes derived from inflated 3D boxes, we adopt an instance-level Gaussian representation similar to GUIDE~\citep{hu2026guide}. Given the lifted 3D center $\bm{c}_{q}$ and decoded feature of each query, we represent its volumetric extent using a compact mixture of $N_g$ anisotropic 3D Gaussians. For each component, lightweight prediction heads estimate a center offset $\Delta\bm{\mu}_{q,g}$, axis-wise standard deviations $\bm{\sigma}_{q,g}$, a rotation matrix $\bm{R}_{q,g}$, and an occupancy weight $\alpha_{q,g}$. Its mean and covariance are defined as

\vspace{-1.em}
\begin{equation}
\bm{\mu}_{q,g} = \bm{c}_{q} + \Delta\bm{\mu}_{q,g},
\qquad
\bm{\Sigma}_{q,g} = \bm{R}_{q,g} \operatorname{diag}\left(\bm{\sigma}_{q,g}^{2}\right) \bm{R}_{q,g}^{\top}.
\end{equation}

The resulting Gaussian components jointly describe the object geometry as a continuous occupancy field that can be discretized at different voxel resolutions. Unlike voxel-grid representations tied to a predefined discretization, our continuous Gaussian representation can be discretized at different resolutions during inference without retraining or modifying the network architecture.

To forecast future occupancy, a lightweight trajectory head predicts a sequence of displacements for each query. These displacements are accumulated from the present 3D center to determine the query location at each future step. The associated Gaussian mixture is then translated along the predicted trajectory while retaining its estimated offsets, covariance, and occupancy weights. This enables the model to generate multiple future occupancy states in a single forward pass.

\subsection{Optimization}
\label{sec:optimization}

In this section, we describe the ground-truth assignment and training objectives used to optimize the instance-wise queries. Since the learned queries are permutation-invariant, we establish a  bipartite correspondence between the predicted queries and ground-truth instances using Hungarian matching~\citep{kuhn1955hungarian}. The matching cost combines the 3D center distance over the observed frames, the spatial overlap between the query attention map and the projected ground-truth instance region, and the predicted instance confidence. Geometric, occupancy, and trajectory supervision is applied only to matched pairs, while unmatched queries are treated as background for confidence supervision.

\begin{table}[t]
    \centering
    \caption{
        Quantitative comparison of dense voxel-wise and instance-wise methods for GMO occupancy forecasting. LighTROcc$^\dagger$ denotes the resolution-matched variant using $704\times256$ inputs. Best and second-best results are shown in bold and underlined, respectively.
    }
    \label{tab:main_results}
    \renewcommand{\arraystretch}{1.15}
    \setlength{\tabcolsep}{4.2pt}

    \resizebox{\linewidth}{!}{
    \begin{tabular}{l c c ccc ccc c}
        \toprule
        Method
        & Type
        & Input Res.
        & $\text{IoU}_{\text{cur}}$
        & $\text{IoU}_{\text{fut}}$
        & $\text{IoU}_{\text{avg}}$
        & $\text{AP}_{\text{cur}}$
        & $\text{AP}_{\text{fut}}$
        & $\text{AP}_{\text{avg}}$
        & FPS \\
        \midrule

        Cam4DOcc~\citep{ma2024cam4docc}
        & \multirow{3}{*}{\makecell{Dense\\Voxel\\wise}}
        & $1600 \times 896$
        & 13.42
        & 11.80
        & 12.12
        & 10.71
        & 8.02
        & 8.56
        & 0.52 \\

        EfficientOCF~\citep{xu2025efficientocf}
        &
        & $1600 \times 896$
        & 8.67
        & 8.59
        & 8.61
        & 14.25
        & 14.13
        & 14.15
        & 3.40 \\

        OccProphet~\citep{chen2025occprophet}
        &
        & $800 \times 448$
        & 17.57
        & 14.23
        & 14.89
        & 20.14
        & 7.25
        & 9.83
        & 3.27 \\

        \midrule

        SparseOcc~\citep{liu2024sparseocc}
        & \multirow{4}{*}{\makecell{Instance\\wise}}
        & $704 \times 256$
        & 18.42
        & 13.67
        & 14.62
        & 20.05
        & 13.36
        & 14.70
        & \underline{14.18} \\

        GUIDE~\citep{hu2026guide}
        &
        & $704 \times 256$
        & 15.21
        & 12.40
        & 12.96
        & 34.73
        & 24.51
        & 26.55
        & 8.16 \\

        \textbf{LighTROcc$^\dagger$~(Ours)}
        &
        & $704 \times 256$
        & \underline{18.46}
        & \underline{14.34}
        & \underline{15.16}
        & \underline{40.03}
        & \underline{31.30}
        & \underline{33.05}
        & \textbf{25.10} \\

        \textbf{LighTROcc~(Ours)}
        &
        & $1600 \times 896$
        & \textbf{19.76}
        & \textbf{15.30}
        & \textbf{16.19}
        & \textbf{42.62}
        & \textbf{33.34}
        & \textbf{35.20}
        & 12.80 \\

        \bottomrule
        \vspace{-2.5em}
    \end{tabular}
    }
\end{table}

The confidence head predicts a foreground probability $s_q \in [0,1]$ for each query. Based on the matching result, matched and unmatched queries are assigned foreground and background targets, respectively. The confidence loss $\mathcal{L}_{\mathrm{conf}}$ is then computed using cross-entropy between the predicted probabilities and the corresponding binary targets. At inference, only queries with $s_q > \tau_{\mathrm{conf}}$ are retained for occupancy prediction, where $\tau_{\mathrm{conf}} = 0.9$.

For each matched pair, we compute the lifted 3D center loss $\mathcal{L}_{\mathrm{cen}}$ using the $\ell_1$ norm and supervise the predicted depth distribution with the cross-entropy loss $\mathcal{L}_{\mathrm{depth}}$. To encourage accurate attention-based 2D localization, the attention localization loss $\mathcal{L}_{\mathrm{attn}}$ is defined as the negative log-likelihood of the normalized attention mass assigned to the projected ground-truth region. To supervise object geometry, the predicted Gaussian mixture is voxelized and optimized against the ground-truth occupancy using focal and Tversky losses~\citep{lin2017focal, salehi2017tversky}, denoted by $\mathcal{L}_{\mathrm{occ}}$. Finally, the trajectory head is supervised with an $\ell_1$ displacement loss $\mathcal{L}_{\mathrm{traj}}$ over the valid future frames. The overall training objective is

\vspace{-0.8em}
\begin{equation}
\mathcal{L}
=
\lambda_{\mathrm{conf}}\mathcal{L}_{\mathrm{conf}}+\lambda_{\mathrm{cen}}\mathcal{L}_{\mathrm{cen}}+\lambda_{\mathrm{depth}}\mathcal{L}_{\mathrm{depth}}+\lambda_{\mathrm{attn}}\mathcal{L}_{\mathrm{attn}}+\lambda_{\mathrm{occ}}\mathcal{L}_{\mathrm{occ}}+\lambda_{\mathrm{traj}}\mathcal{L}_{\mathrm{traj}},
\end{equation}

where each $\lambda$ controls the contribution of the corresponding loss term.


\section{Experiments}
\label{sec:experiment}

\subsection{Experimental Setup}
\label{sec:experimental_setup}

\paragraph{Datasets.}
We train and evaluate LighTROcc and the baseline models using nuScenes~\citep{caesar2020nuscenes} and nuScenes-Occupancy~\citep{wang2023openoccupancy}. We obtain multi-view RGB images, camera parameters, and ego poses from nuScenes, and use the fine-grained voxel occupancy annotations from nuScenes-Occupancy for supervision. Following the prior work~\citep{ma2024cam4docc, xu2025efficientocf, chen2025occprophet}, we use 700 scenes for training and 150 for evaluation, yielding 23,930 and 5,119 sequences, respectively. Each sequence contains two past observations, one present observation, and four future occupancy targets. Since the nuScenes keyframes are sampled at 2 Hz, these targets correspond to prediction horizons of 0.5, 1.0, 1.5, and 2.0 seconds. The evaluation range is $[-51.2, 51.2]$ m along both horizontal axes and $[-5, 3]$ m vertically, discretized into a $[512 \times 512 \times 40]$ grid with a voxel size of $0.2$ m. Since the LiDAR-derived nuScenes-Occupancy annotations are highly sparse, we supplement them with auxiliary assets. Further details are provided in the Appendix.

\paragraph{Evaluation protocols and metrics.}
Following prior works~\citep{ma2024cam4docc, xu2025efficientocf, chen2025occprophet}, we evaluate foreground general movable object (GMO) instances in both the present and future frames. For current-only methods, we follow Cam4DOcc and EfficientOCF by using current-frame predictions for all future steps. To assess occupancy forecasting performance, we report voxel-wise IoU against the fine-grained 3D occupancy ground truth. For instance-level evaluation, we additionally report average precision (AP) based on the voxel IoU between predicted and ground-truth instance occupancy masks. Following GUIDE~\citep{hu2026guide}, we compute AP at IoU thresholds of 0.1, 0.2, and 0.3 and report the mean. For dense voxel-wise methods, we recover instance masks using the official flow-based grouping and compute AP using an evaluation scheme inspired by Panoptic-DeepLab~\citep{cheng2020panoptic}. For center localization, we report $\mathrm{AP}_{\mathrm{ctr}}$ and $\mathrm{Recall}_{\mathrm{loc}}$ at a $1$ m threshold, measuring confidence-ranked center localization and active-query GT coverage, respectively. All baseline models are reproduced using their official code, and all methods are evaluated on a single NVIDIA RTX Pro 6000 GPU. Further details are provided in the Appendix.

\subsection{Experimental Results}
\label{sec:experimental_results}

\begin{figure}[t]
    \centering
    \includegraphics[width=\textwidth]{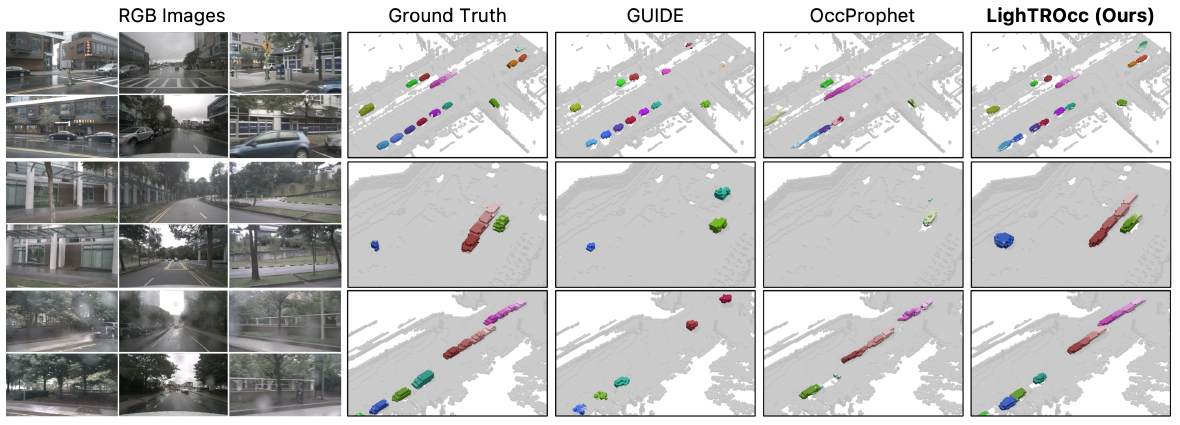}
    \vspace{-2.0em}
    \caption{
        Qualitative comparison of GMO occupancy predictions from multi-view RGB images. LighTROcc more faithfully captures object locations and future motion, with darker colors indicating later forecasting steps.
    }
    \vspace{-1.2em}
    \label{fig:qualitative}
\end{figure}


\paragraph{Forecasting General Movable Object Occupancy.}

We conduct quantitative comparisons to evaluate 4D occupancy forecasting for general movable objects. For a resolution-matched comparison with prior instance-wise methods, we additionally report LighTROcc$^\dagger$, a lighter variant using $704 \times 256$ inputs. As shown in Table~\ref{tab:main_results}, dense voxel-wise approaches achieve competitive voxel-wise IoU, indicating effective reconstruction of overall occupied regions. However, their instance-agnostic forecasts require flow-based post-processing to recover object masks, resulting in relatively low AP and weaker object-level consistency over time. In contrast, instance-wise approaches generally achieve higher AP while maintaining competitive IoU. LighTROcc obtains the best results across all reported IoU and AP metrics, reaching $\text{IoU}_{\text{avg}}=16.19$ and $\text{AP}_{\text{avg}}=35.20$ while running at $12.80$ FPS. Notably, LighTROcc$^\dagger$ retains comparable occupancy accuracy while running nearly twice as fast as the full-resolution model. These results demonstrate that LighTROcc maintains reliable occupancy forecasting and object-level consistency across different input resolutions while remaining suitable for real-time inference.

\paragraph{Qualitative results.}
To qualitatively assess LighTROcc, we compare its predictions with representative baselines. As shown in Figure~\ref{fig:qualitative}, the results visualize instance-wise GMO occupancy and its evolution over time, while GUIDE~\citep{hu2026guide} is shown only for the current frame because it does not model future states. GUIDE localizes objects well in the current frame, but cannot forecast future occupancy due to the absence of future-state modeling. OccProphet~\citep{chen2025occprophet} supports future occupancy forecasting, but often fails to preserve consistent object geometry across time. In contrast, LighTROcc more accurately captures future object locations while maintaining consistent object representations throughout the forecasting horizon. Its accurate trajectory forecasting further demonstrates its potential for downstream planning and control in autonomous driving. Additional qualitative results and analyses are provided in the Appendix.

\subsection{Computational Efficiency and Model Compactness}
\label{sec:efficiency}
We compare LighTROcc with prior baselines in terms of computational efficiency and model compactness. Table~\ref{tab:efficiency} reports parameters, GFLOPs, latency, FPS, and forecasting accuracy, with all models benchmarked using official implementations on a single NVIDIA RTX Pro 6000 GPU. Among current-only methods, SparseOcc~\citep{liu2024sparseocc} is the most efficient, partly aided by an input resolution nearly $8\times$ lower than our full-resolution setting. Among 4D forecasting methods, LighTROcc achieves both the highest inference speed and the best accuracy. For a resolution-matched comparison, LighTROcc$^{\dagger}$ runs about $1.8\times$ faster than SparseOcc while also achieving higher accuracy. Notably, the full-resolution LighTROcc maintains real-time inference despite processing high-resolution inputs and forecasting multiple frames in a single pass, while achieving the best performance. This highlights the effectiveness of our compact instance-centric formulation.

\begin{table}[t]
    \centering
    \caption{
        Comparison of model compactness, computational efficiency, and forecasting accuracy.
        Runtime is measured using official implementations on a single NVIDIA RTX Pro 6000 GPU.
        LighTROcc$^\dagger$ denotes the resolution-matched variant using $704\times256$ inputs.
        Best and second-best results are shown in bold and underlined, respectively.
    }
    \label{tab:efficiency}

    \renewcommand{\arraystretch}{1.15}
    \setlength{\tabcolsep}{4.5pt}

    \resizebox{\linewidth}{!}{
    \begin{tabular}{c l c c c c c c c}
        \toprule
        Category
        & Method
        & \makecell{Input Res.}
        & \makecell{Params.\\(M)}
        & GFLOPs
        & \makecell{Latency\\(ms)}
        & FPS
        & $\text{IoU}_{\text{avg}}$
        & $\text{AP}_{\text{avg}}$ \\
        \midrule

        \multirow[c]{2}{*}[-2pt]{Current-only}
        & SparseOcc~\citep{liu2024sparseocc}
        & $704 \times 256$
        & 80
        & \underline{568}
        & \underline{70.53}
        & \underline{14.18}
        & 14.62
        & 14.70 \\

        \cmidrule(lr){2-9}

        & GUIDE~\citep{hu2026guide}
        & $704 \times 256$
        & 90
        & 1,530
        & 122.51
        & 8.16
        & 12.96
        & 26.55 \\

        \midrule

        \multirow[c]{5}{*}[-7pt]{\makecell{4D\\forecasting}}
        & Cam4DOcc~\citep{ma2024cam4docc}
        & $1600 \times 896$
        & 370
        & 13,168
        & 1,902.26
        & 0.52
        & 12.12
        & 8.56 \\

        \cmidrule(lr){2-9}

        & EfficientOCF~\citep{xu2025efficientocf}
        & $1600 \times 896$
        & 388
        & 5,143
        & 294.24
        & 3.40
        & 8.61
        & 14.15 \\

        \cmidrule(lr){2-9}

        & OccProphet~\citep{chen2025occprophet}
        & $800 \times 448$
        & 82
        & 2,020
        & 305.93
        & 3.27
        & 14.89
        & 9.83 \\

        \cmidrule(lr){2-9}

        & \textbf{LighTROcc$^\dagger$~(Ours)}
        & $704 \times 256$
        & \textbf{42}
        & \textbf{154}
        & \textbf{39.84}
        & \textbf{25.10}
        & \underline{15.16}
        & \underline{33.05} \\

        \cmidrule(lr){2-9}

        & \textbf{LighTROcc~(Ours)}
        & $1600 \times 896$
        & \textbf{42}
        & 1,225
        & 78.12
        & 12.80
        & \textbf{16.19}
        & \textbf{35.20} \\

        \bottomrule
    \end{tabular}
    }
\end{table}

\subsection{Ablation Studies}
\label{sec:ablation}

\begin{figure}[t]
    \centering
    \includegraphics[width=\textwidth]{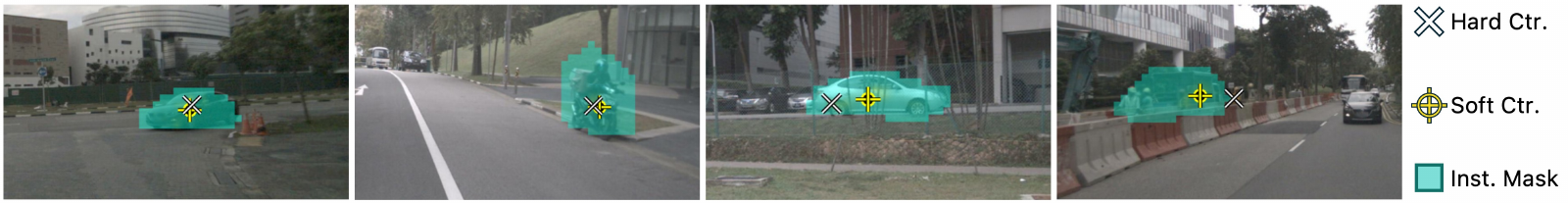}
    \vspace{-2.0em}
    \caption{
        Qualitative comparison of hard- and soft-argmax localization from query attention maps. Hard argmax is sensitive to spurious outlier peaks, while soft argmax provides more reliable image-plane centers within the corresponding instance regions.
    }
    \label{fig:attention_map}
\end{figure}

\paragraph{Ablation on 3D center localization.}
\begin{table}[t]
    \centering
    \vspace{-1.1em}
    \begin{minipage}[t]{0.58\linewidth}
        \vspace{0pt}
        \centering

        \caption{Ablation of 3D center localization strategies. Direct 3D regression is compared with hard- and soft-argmax variants of attention-guided forward lifting under the same model configuration.}
        \label{tab:localization_ablation}

        \fontsize{7.5pt}{8.5pt}\selectfont
        \setlength{\tabcolsep}{2.5pt}
        \renewcommand{\arraystretch}{1.1}
        \setlength{\aboverulesep}{0.35ex}
        \setlength{\belowrulesep}{0.35ex}

        \begin{tabularx}{\linewidth}{
            @{}
            >{\raggedright\arraybackslash}X
            >{\centering\arraybackslash}p{0.14\linewidth}
            >{\centering\arraybackslash}p{0.14\linewidth}
            >{\centering\arraybackslash}p{0.13\linewidth}
            >{\centering\arraybackslash}p{0.17\linewidth}
            @{}
        }
            \toprule
            Localization
            & $\mathrm{IoU}_{\mathrm{avg}}\!\uparrow$
            & $\mathrm{AP}_{\mathrm{avg}}\!\uparrow$
            & $\mathrm{AP}_{\mathrm{ctr}}\!\uparrow$
            & $\mathrm{Recall}_{\mathrm{loc}}\!\uparrow$ \\
            \midrule

            Direct regression (MLP)
            & 11.13
            & 26.10
            & 17.23
            & 21.14 \\

            Forward Lifting (hard)
            & 11.20
            & 28.50
            & 25.43
            & 29.84 \\

            \textbf{Forward Lifting (soft)}
            & \textbf{16.19}
            & \textbf{35.20}
            & \textbf{27.31}
            & \textbf{31.06} \\

            \bottomrule
        \end{tabularx}
    \end{minipage}
    \hfill
    \begin{minipage}[t]{0.40\linewidth}
        \vspace{0pt}
        \centering

        \caption{Effect of instance query count on computational efficiency and forecasting accuracy. All variants are evaluated using full-resolution inputs.}
        \label{tab:query_ablation}

        \fontsize{7.5pt}{8.5pt}\selectfont
        \setlength{\tabcolsep}{2.5pt}
        \renewcommand{\arraystretch}{1.1}
        \setlength{\aboverulesep}{0.35ex}
        \setlength{\belowrulesep}{0.35ex}

        \begin{tabularx}{\linewidth}{
            @{}
            >{\centering\arraybackslash}X
            >{\centering\arraybackslash}X
            >{\centering\arraybackslash}X
            >{\centering\arraybackslash}X
            @{}
        }
            \toprule
            \# Queries
            & FPS $\uparrow$
            & $\mathrm{IoU}_{\mathrm{avg}} \uparrow$
            & $\mathrm{AP}_{\mathrm{avg}} \uparrow$ \\
            \midrule

            100
            & \textbf{12.95}
            & 15.16
            & 34.01 \\

            \textbf{200}
            & 12.80
            & \textbf{16.19}
            & \textbf{35.20} \\

            900
            & 11.42
            & 14.13
            & 31.61 \\

            \bottomrule
        \end{tabularx}
    \end{minipage}

    \vspace{-1.2em}
\end{table}
A key factor enabling accurate instance recognition and localization with a compact query set is the proposed attention-guided forward lifting. Because decoded queries primarily encode 2D visual evidence, directly regressing metric 3D centers without explicit spatial priors is highly underconstrained. Forward lifting eases this localization problem, and we further employ soft argmax rather than hard peak selection to obtain stable image-plane centers. To isolate the contribution of each design choice, we compare direct MLP regression, hard-argmax lifting, and our soft-argmax lifting. As shown in Table~\ref{tab:localization_ablation}, direct regression yields the lowest $\mathrm{IoU}_{\mathrm{avg}}$ and $\mathrm{AP}_{\mathrm{avg}}$, together with substantially lower $\mathrm{AP}_{\mathrm{ctr}}$ and $\mathrm{Recall}_{\mathrm{loc}}$, indicating that poor 3D grounding also degrades overall occupancy prediction. Hard-argmax lifting improves both forecasting accuracy and center localization, but remains vulnerable to spurious outlier peaks in the attention map. In contrast, soft argmax aggregates the full attention distribution, providing more reliable center grounding and achieving the best performance across all four metrics. Figure~\ref{fig:attention_map} further shows that the learned attention maps capture the corresponding object regions, while soft argmax estimates their centers more reliably than hard argmax.

\vspace{-1.0em}
\paragraph{Query scaling and efficiency-accuracy trade-off.}
We analyze how query count affects instance-centric perception and forecasting. Many prior approaches based on fixed spatial anchors initialize 900 or more query candidates to ensure sufficient scene coverage~\citep{lin2023sparse4dv2,liu2023sparsebev, hu2026guide}. In contrast, our attention-guided forward lifting explicitly localizes queries, allowing LighTROcc to operate with a substantially smaller query set. Table~\ref{tab:query_ablation} compares the resulting accuracy and computational overhead, with all variants trained using full-resolution inputs.

Empirically, 200 queries provide the best balance between accuracy and efficiency. Reducing the query count to 100 yields only a marginal speed improvement while slightly degrading AP and IoU, indicating insufficient representation capacity. Conversely, increasing the query count by $4.5\times$ to 900, comparable to the scale commonly used in anchor-based methods, introduces substantial attention overhead and noticeably degrades prediction accuracy. This suggests that excessive query redundancy is not only computationally inefficient but can also hinder effective instance prediction. We provide further statistical analysis of query scaling below. We therefore adopt 200 queries as our default setting, which achieves the most favorable balance between accuracy and efficiency.

\begin{figure}[t]
    \centering
    \includegraphics[width=\textwidth]{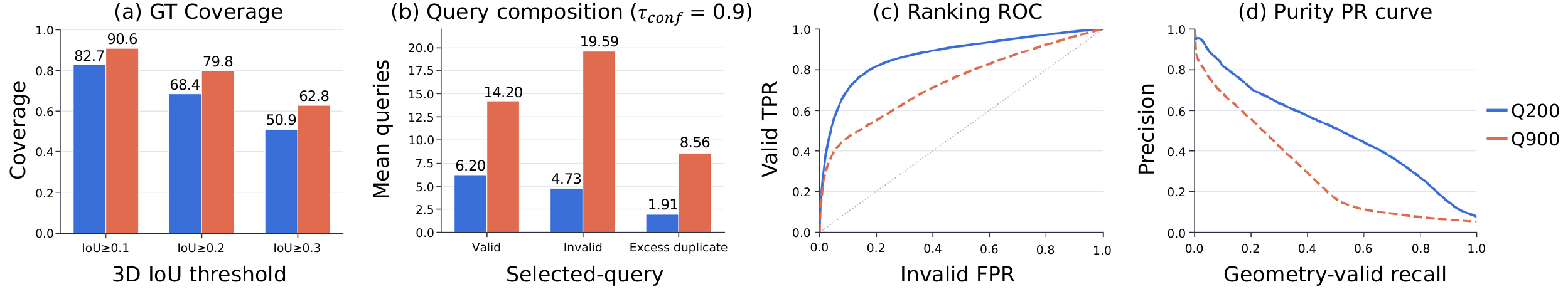}
    \vspace{-1.5em}
    \caption{
        Statistical analysis of query scaling over the evaluation set. Queries with voxel IoU $\geq 0.2$ are treated as geometry-valid. Although 900 queries improve GT coverage, they introduce more redundant and invalid predictions, whereas 200 queries provide stronger confidence discrimination and cleaner query selection.
    }
    \vspace{-1.5em}
    \label{fig:query-performace}
\end{figure}

\paragraph{Statistical analysis of query scaling.}
We further analyze the effect of query scaling over the entire evaluation set. For this analysis, we regard a query as geometry-valid if its voxel IoU with a GT object is at least 0.2, and invalid otherwise. As shown in Figure~\ref{fig:query-performace}~(a), increasing the number of queries from 200 to 900 consistently improves GT object coverage across different voxel-IoU thresholds. However, this increased coverage comes with substantial redundancy. Figure~\ref{fig:query-performace}~(b) shows that the number of invalid queries grows considerably faster than the number of valid queries, while excess duplicate predictions also increase markedly. This excessive redundancy degrades the model's discriminative ability, making it harder to separate valid queries from redundant or invalid predictions based on confidence.

This behavior is reflected in the confidence statistics. In Figure~\ref{fig:query-performace}~(c), the 200-query model exhibits a consistently stronger ROC curve, indicating better discrimination between valid and invalid predictions. Figure~\ref{fig:query-performace}~(d) provides a complementary view: across confidence thresholds, the 200-query model captures a larger fraction of geometry-valid queries while admitting fewer invalid ones. Together, these results indicate that although additional queries improve scene coverage, excessive query redundancy weakens confidence-based selection and degrades instance-level prediction quality. This analysis supports our choice of 200 queries as a compact yet expressive query set.


\section{Conclusion}
\label{sec:conclusion}
\vspace{-0.4em}

We presented LighTROcc, a lightweight instance-centric framework for camera-based 4D occupancy forecasting. LighTROcc represents movable objects with a compact set of learned queries, avoiding dense spatiotemporal feature processing and large sets of spatially initialized candidates. Each query is localized through attention-guided forward lifting, modeled with a mixture of 3D Gaussians, and propagated to future frames using predicted displacements. This design predicts present and future occupancy in a single forward pass while maintaining instance-specific representations. Experiments on nuScenes and supplemented nuScenes-Occupancy show that LighTROcc outperforms all evaluated baselines in IoU and AP while achieving the highest inference speed among 4D forecasting methods. Ablations validate the localization and compact query design, demonstrating an accuracy-efficiency trade-off for real-time 4D occupancy forecasting.

\subsection*{AI use statement}

We used AI in a limited manner for grammar and style polishing only. It did not contribute to
research ideation, experimental design, or analysis. The authors take full responsibility for all
content.

\subsection*{Reproducibility statement}

We provide the model architecture and training objectives in Section~\ref{sec:method}, and
the datasets and evaluation protocols in Section ~\ref{sec:experimental_setup}.
Appendix~\ref{sec:appendix_data_and_eval} details the dataset construction and evaluation metrics, while
Appendix~\ref{sec:implementation_details} provides the network architecture and training implementation details.



\bibliography{iclr2027_conference}
\bibliographystyle{iclr2027_conference}

\clearpage
\appendix

\section{Evaluation Protocol and Dataset Analysis}
\label{sec:appendix_data_and_eval}

\subsection{Evaluation Metrics}
\label{sec:appendix_metrics}

\paragraph{Occupancy and instance evaluation.}
In this section, we provide additional details on the evaluation protocol for the instance-level average precision (AP) and voxel-wise intersection over union (IoU) reported in the main paper. For instance-level AP, each predicted object is represented by a voxel occupancy mask and an instance confidence score. Predicted instances are matched to ground-truth instances based on their voxel IoU. A prediction is considered a true positive if its IoU with the matched ground-truth instance exceeds a threshold $\tau$, while unmatched predictions are counted as false positives. We compute AP independently at $\tau \in \{0.1, 0.2, 0.3\}$ and report the mean across these thresholds for each temporal step. 

Let $\mathrm{AP}_k$ denote the threshold-averaged AP at time step $k$, where $k=0$ corresponds to the present frame and $k\in\{1,\ldots,T_f\}$ denotes a future frame. The reported temporal metrics are computed as
\begin{equation}
    \mathrm{AP}_{\mathrm{cur}} = \mathrm{AP}_0,
    \qquad
    \mathrm{AP}_{\mathrm{fut}}
    = \frac{1}{T_f}\sum_{k=1}^{T_f}\mathrm{AP}_k,
\end{equation}
and
\begin{equation}
    \mathrm{AP}_{\mathrm{avg}}
    = \frac{1}{T_f+1}\sum_{k=0}^{T_f}\mathrm{AP}_k.
\end{equation}
Thus, $\mathrm{AP}_{\mathrm{fut}}$ measures forecasting accuracy exclusively 
over future frames, whereas $\mathrm{AP}_{\mathrm{avg}}$ summarizes performance 
over both the present and future states.

For voxel-wise evaluation, all GMO instances at each temporal step are merged 
into a binary foreground occupancy map. Given the predicted and ground-truth 
foreground voxel sets $\mathcal{P}_k$ and $\mathcal{G}_k$, respectively, we 
compute
\begin{equation}
    \mathrm{IoU}_k
    =
    \frac{|\mathcal{P}_k \cap \mathcal{G}_k|}
         {|\mathcal{P}_k \cup \mathcal{G}_k|}.
\end{equation}
$\mathrm{IoU}_{\mathrm{cur}}$, $\mathrm{IoU}_{\mathrm{fut}}$, and 
$\mathrm{IoU}_{\mathrm{avg}}$ are aggregated over the temporal dimension in 
the same manner as their AP counterparts.

For dense voxel-wise baselines, instance masks required for AP evaluation are 
recovered from their occupancy and flow predictions using the corresponding 
official grouping procedure. This conversion is used only for instance-level 
evaluation, while voxel-wise IoU is computed directly from the predicted 
foreground occupancy.

\paragraph{Center localization evaluation.}
In addition to occupancy prediction, we evaluate query-level center localization on the current frame. We report $\mathrm{AP}_{\mathrm{ctr}}$ and $\mathrm{Recall}_{\mathrm{loc}}$ at a $1$ m threshold, measuring confidence-ranked center localization and active-query GT coverage, respectively. For $\mathrm{AP}_{\mathrm{ctr}}$, foreground queries are matched to ground-truth instances based on the center distance. Queries are processed in descending confidence order, and a prediction is considered a true positive if its distance to an unmatched ground-truth instance is within the threshold. The resulting precision-recall curve is used to compute $\mathrm{AP}_{\mathrm{ctr}}$. For $\mathrm{Recall}_{\mathrm{loc}}$, we evaluate active queries after inference-time filtering and match them with ground-truth instances using one-to-one radius matching. It measures the ratio of successfully localized ground-truth instances:
\begin{equation}
    \mathrm{Recall}_{\mathrm{loc},\tau}
    =
    \frac{N_{\mathrm{matched}}^{d\leq\tau}}
         {N_{\mathrm{GT}}},
\end{equation}
where $N_{\mathrm{matched}}^{d\leq\tau}$ denotes the number of ground-truth instances matched within the distance threshold $\tau$. Both metrics are computed using center distances.

\subsection{Dataset Analysis and Ground Truth Supplementation}
\label{sec:appendix_gt_supplement}

\begin{figure}[h]
    \centering
    \includegraphics[width=\textwidth]{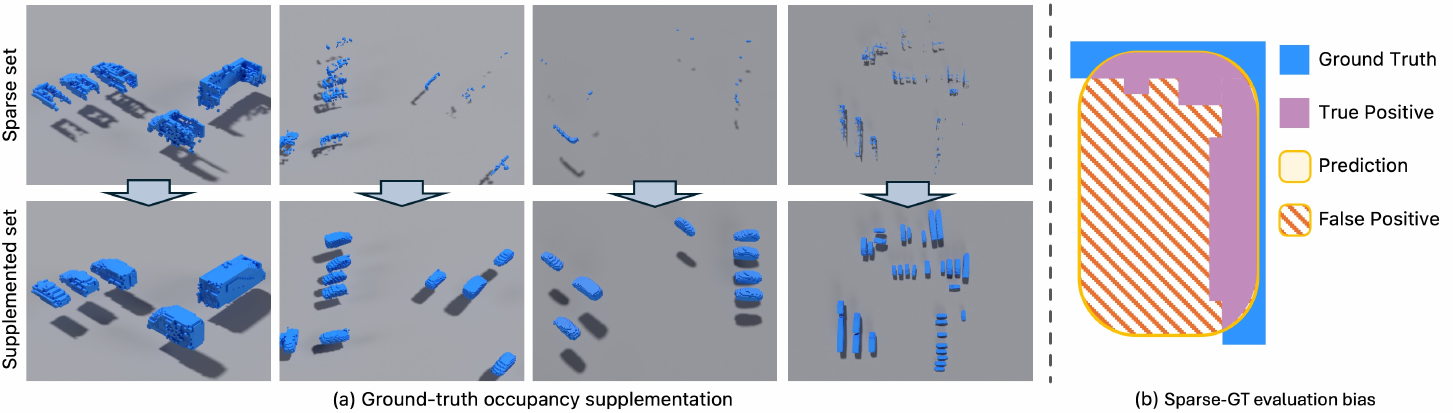}
    \vspace{-2.0em}
    \caption{
        Illustration of ground-truth occupancy supplementation and evaluation bias from sparse annotations. (a) Original sparse occupancy annotations (top) and their supplemented counterparts using 3D assets (bottom). (b) Sparse ground truth can introduce apparent false-positive predictions even when the predicted geometry is physically reasonable.
    }
    \label{fig:appendix_asset}
\end{figure}

Following prior works~\citep{ma2024cam4docc, xu2025efficientocf, chen2025occprophet}, we evaluate occupancy predictions using the ground-truth annotations from nuScenes-Occupancy~\citep{wang2023openoccupancy}. Since its annotations are derived from LiDAR observations, however, they are inherently incomplete. LiDAR primarily captures visible object surfaces rather than their complete volumetric extent, often resulting in hollow or incomplete occupancy structures even when the overall object shape appears visually plausible. This issue becomes more pronounced for distant or occluded objects, where only a small fraction of the object surface may be observed and the resulting occupancy can be severely fragmented. The upper row of Figure~\ref{fig:appendix_asset}~(a) illustrates representative examples of such sparse annotations. Beyond their limited visual completeness, these annotations can also introduce undesirable biases in both training and evaluation.

Nearby objects generally contain denser occupancy observations and therefore provide relatively strong signal. In contrast, distant or low-visibility objects may occupy only a few voxels, providing substantially weaker signals. This imbalance can bias a model toward predicting well-observed nearby objects while suppressing predictions for difficult instances. As illustrated in Figure~\ref{fig:appendix_asset}~(b), predicting the full spatial extent of a sparsely annotated object can introduce many apparent false-positive voxels, even when the predicted geometry is physically reasonable. Conversely, missing sparsely observed regions may incur only a limited penalty under voxel-based evaluation. This creates an unfavorable evaluation bias that can encourage prediction suppression for difficult objects. Such behavior is undesirable in autonomous driving, where reliably detecting distant and partially occluded objects is particularly important.

To mitigate this issue, we supplement the original sparse occupancy annotations with 3D geometric prototypes from ShapeNet~\citep{Chang:2015:SAI} and SMPL~\citep{SMPL:2015}. For each object, we obtain its 3D bounding-box center, dimensions, orientation, and semantic category from the nuScenes annotations, and place a corresponding 3D mesh according to its estimated pose and extent. We use watertight meshes so that the enclosed volume can be treated as occupied rather than retaining only the visible surface. Importantly, the original nuScenes-Occupancy annotation is preserved rather than replaced. The asset-based occupancy is combined with the original annotation through a voxel-wise union, thereby retaining the observed object geometry while completing severely missing regions. The lower row of Figure~\ref{fig:appendix_asset}~(a) visualizes the resulting supplemented annotations. This procedure provides more balanced evaluation across objects with different distances and visibility levels, while imposing a clearer penalty for missed instances and reducing the artificial false-positive penalty caused by sparse LiDAR observations.

To quantify the effectiveness of supplementation and its impact on evaluation fairness, we measure voxel density under different distance and visibility conditions. We analyze 518,702 3D object annotations from 29,049 training and evaluation sequences. The voxel density is computed as

\begin{equation}
    D = \frac{\sum_i |S_i \cap B_i|}{\sum_i |B_i|},
\end{equation}

where $S_i$ denotes either the original sparse voxel set or the supplemented voxel set of object $i$, and $B_i$ denotes the set of voxels inside its ground-truth 3D bounding box from nuScenes. As shown in Table~\ref{tab:occupancy_density}, the overall density
increases from 17.62\% to 71.51\% ($4.06\times$).

When grouped by distance, the original annotations become increasingly sparse as objects move farther away, with density falling below 14\% beyond 30~m and reaching only 8.11\% beyond 50~m. Consequently, even predictions that correctly cover the object location and extent can incur a large number of apparent false-positive voxels under the original annotations. In contrast, the supplemented annotations maintain substantially more stable densities across distance, ranging only from 74.06\% to 68.90\%, thereby providing more consistent evaluation for distant objects. A similar trend is observed across visibility levels, where supplementation substantially reduces the density gap between poorly and highly visible objects. These results show that the supplemented annotations effectively mitigate the distance- and visibility-dependent sparsity of the original annotations, reducing the resulting evaluation bias and enabling a more balanced assessment across different observation conditions.


\begin{table}[h]
    \centering
    \caption{
        Occupancy density before and after 3D asset supplementation. Gain denotes the ratio of supplemented to original occupancy density.
    }
    \label{tab:occupancy_density}
    \small
    \setlength{\tabcolsep}{8pt}
    \renewcommand{\arraystretch}{1.08}

    \begin{tabular}{llccc}
        \toprule
        Subset
        & Range
        & Original (\%)
        & Supplemented (\%)
        & Gain ($\times$) \\
        \midrule

        Overall
        & All
        & 17.62
        & \textbf{71.51}
        & \textbf{4.06} \\

        \midrule

        \multirow{6}{*}{Distance (m)}
        & $[0,10)$
        & 28.07
        & 74.06
        & 2.64 \\

        & $[10,20)$
        & 23.39
        & 72.73
        & 3.11 \\

        & $[20,30)$
        & 17.63
        & 71.59
        & 4.06 \\

        & $[30,40)$
        & 13.95
        & 70.79
        & 5.08 \\

        & $[40,50)$
        & 11.55
        & 69.97
        & 6.06 \\

        & $\geq 50$
        & 8.11
        & 68.90
        & 8.50 \\

        \midrule

        \multirow{4}{*}{Visibility (\%)}
        & V1 (0 to 40)
        & 11.63
        & 69.19
        & 5.95 \\

        & V2 (40 to 60)
        & 14.54
        & 70.24
        & 4.83 \\

        & V3 (60 to 80)
        & 16.94
        & 71.15
        & 4.20 \\

        & V4 (80 to 100)
        & 21.94
        & 73.24
        & 3.34 \\

        \bottomrule
    \end{tabular}
\end{table}

\subsection{Geometric Consistency of Supplemented Annotations}
\label{sec:appendix_geometric_consistency}

In this section, we perform complementary geometric checks to validate the supplemented annotations against the original nuScenes-Occupancy~\citep{wang2023openoccupancy} annotations and LiDAR observations. We first measure the surface distance to quantify how closely the added assets fit the original LiDAR-derived occupied voxels. For each annotated object, let $\mathcal{S}_i$ denote the centers of its original occupied voxels and $\mathcal{M}_i$ the surface of the fitted asset. For each $\mathbf{x}\in\mathcal{S}_i$, we compute the nearest-surface distance as

\begin{equation}
d(\mathbf{x},\mathcal{M}_i)
=
\min_{\mathbf{y}\in\mathcal{M}_i}
\|\mathbf{x}-\mathbf{y}\|_2.
\end{equation}

We report the mean and 90th-percentile distances, where smaller values indicate closer agreement with the geometry captured by the original observations.

We additionally measure the free-space violation rate to assess whether the supplemented geometry introduces excessive occupancy beyond the observed object geometry. We trace the original LiDAR rays and define the observed free space along each ray up to its measured return range as

\begin{equation}
\mathcal{F}(r)
=
\left\{
\mathbf{s}+t\mathbf{d}
\;\middle|\;
0 \leq t < r
\right\},
\end{equation}

where $\mathbf{s}$ is the LiDAR sensor origin, $\mathbf{d}$ is the unit ray direction, $r$ is the measured return range, and $t$ is the distance from the sensor origin along the ray. We then measure the fraction of supplemented occupancy that overlaps this observed free space, where a lower value indicates less conflict with the original observations.

As shown in Table~\ref{tab:gt_geometry_validation}, the fitted assets achieve a mean nearest-surface distance of 0.20\,m and a $P90$ distance of 0.38\,m. Given the 0.2\,m voxel resolution used for evaluation, these correspond to approximately 1.0 and 1.9 voxels, respectively. The free-space violation rate is only 2.4\%, indicating that the supplemented geometry rarely conflicts with directly observed free space. Together with the density analysis in Appendix~\ref{sec:appendix_gt_supplement}, these results show that the supplementation increases geometric completeness while remaining largely consistent with the original sensor evidence.

\begin{table}[t]
\centering
\caption{
Geometric consistency of supplemented annotations.
Surface distance measures agreement with occupied voxels, while free-space violation measures overlap with LiDAR-observed free space.
}
\label{tab:gt_geometry_validation}
\begin{tabular}{lc}
\toprule
Metric & Result \\
\midrule
Nearest-surface distance, mean $\downarrow$ & 0.20 m \\
Nearest-surface distance, $P90$ $\downarrow$ & 0.38 m \\
LiDAR free-space violation $\downarrow$ & 2.4\% \\
\bottomrule
\end{tabular}
\end{table}

\subsection{Robustness to Ground-Truth Annotation Protocol}
\label{sec:gt_protocol_robustness}

To examine whether the comparative advantage of LighTROcc depends on the supplemented annotations, we additionally evaluate all methods using the original sparse nuScenes-Occupancy annotations. As analyzed in Appendix~\ref{sec:appendix_gt_supplement}, these annotations cover only 17.62\% of the object bounding-box volume on average and become increasingly sparse with distance. Consequently, even geometrically plausible predictions may be counted as false positives when they occupy regions that are simply unobserved in the sparse ground truth.

To mitigate this annotation-induced penalty, we adopt a conservative bbox-aware evaluation protocol. Predicted voxels that lie inside annotated 3D bounding boxes but are absent from the sparse ground truth are excluded from false-positive counting, without being reassigned as true positives. The resulting IoU is computed as

\begin{equation}
\operatorname{IoU}
=
\frac{TP}
{TP + FN + FP - FP^{\mathrm{bbox}}},
\end{equation}

where $FP^{\mathrm{bbox}}$ denotes false-positive voxels inside the annotated 3D bounding boxes. The same ignore mask is applied when computing voxel overlaps for instance-level AP. Thus, the protocol does not assume that the entire bounding box is occupied, but only avoids treating unlabeled in-box regions as reliable negative evidence.

As shown in Table~\ref{tab:original_sparse_eval}, LighTROcc retains the highest aggregate performance, achieving $\mathrm{IoU}_{\mathrm{avg}}=9.16$ and $\mathrm{AP}_{\mathrm{avg}}=13.93$. Although SparseOcc achieves a slightly higher current-frame IoU, LighTROcc obtains the best future and average IoU and the best AP across all temporal aggregations. Moreover, Table~\ref{tab:main_results} shows that all evaluated methods achieve higher average IoU and AP under the supplemented annotations than under the original sparse-label protocol. This consistent improvement across methods suggests that supplementation alleviates a common sparsity-induced evaluation penalty rather than providing an advantage specific to LighTROcc.

Together with the density analysis in Appendix~\ref{sec:appendix_gt_supplement}, these results provide complementary evidence for our evaluation design. The original sparse annotations can introduce unintended evaluation bias due to their incomplete object coverage. Although the bbox-aware evaluation mitigates this issue by excluding ambiguous in-box false positives, it still provides no positive target for unobserved object regions. Consequently, missing predictions in these regions are not counted as false negatives, particularly for distant or occluded objects where the annotations are severely incomplete. To address both limitations, we adopt the supplemented annotations as a more complete evaluation target. The quantitative analyses and comparative experiments collectively support the rationale and effectiveness of this supplementation strategy.

\begin{table}[t]
\centering
\caption{
Quantitative comparison of dense voxel-wise and instance-wise methods under the original sparse nuScenes-Occupancy annotations using the bbox-aware evaluation protocol. Best and second-best results are shown in bold and underlined, respectively.
}
\label{tab:original_sparse_eval}
\renewcommand{\arraystretch}{1.15}
\setlength{\tabcolsep}{4.2pt}
\scalebox{0.88}{
\begin{tabular}{l c ccc ccc}
    \toprule
    Method
    & Type
    & $\text{IoU}_{\text{cur}}$
    & $\text{IoU}_{\text{fut}}$
    & $\text{IoU}_{\text{avg}}$
    & $\text{AP}_{\text{cur}}$
    & $\text{AP}_{\text{fut}}$
    & $\text{AP}_{\text{avg}}$ \\
    \midrule

    Cam4DOcc~\citep{ma2024cam4docc}
    & \multirow{3}{*}{\makecell{Dense\\Voxel\\wise}}
    & 11.23
    & 3.84
    & 5.32
    & 8.70
    & 2.80
    & 3.98 \\

    EfficientOCF~\citep{xu2025efficientocf}
    &
    & 1.52
    & 1.43
    & 1.45
    & \underline{11.56}
    & \underline{10.94}
    & \underline{11.06} \\

    OccProphet~\citep{chen2025occprophet}
    &
    & 10.50
    & \underline{8.09}
    & 8.57
    & 11.00
    & 8.13
    & 8.70 \\

    \midrule

    SparseOcc~\citep{liu2024sparseocc}
    & \multirow{3}{*}{\makecell{Instance\\wise}}
    & \textbf{11.91}
    & 8.06
    & \underline{8.83}
    & 9.48
    & 8.14
    & 8.41 \\

    GUIDE~\citep{hu2026guide}
    &
    & 6.48
    & 5.68
    & 5.84
    & 10.71
    & 9.77
    & 9.96 \\

    \textbf{LighTROcc~(Ours)}
    &
    & \underline{11.42}
    & \textbf{8.59}
    & \textbf{9.16}
    & \textbf{17.56}
    & \textbf{13.02}
    & \textbf{13.93} \\

    \bottomrule
    \vspace{-2.5em}
\end{tabular}
}
\end{table}

\section{Implementation Details}
\label{sec:implementation_details}

\subsection{Network Architecture}
\label{sec:appendix_arcitecture}

We use a ResNet-18~\citep{he2016deep} initialized with pretrained weights as the image backbone for feature extraction. The extracted image features are projected to the same embedding dimension as the instance queries, with $D=128$, before being processed by the transformer decoder. We initialize a compact set of 200 learnable instance queries. The transformer decoder consists of $L=3$ layers with $H=4$ attention heads. To encode the spatial, camera-specific, and temporal information of the image tokens, we incorporate camera-parameter embeddings together with 2D sinusoidal positional encodings, learned camera-ID embeddings, and learned temporal embeddings.

The depth head is implemented as a lightweight two-layer multilayer perceptron (MLP) that predicts a categorical depth distribution over $N_d=64$ bins within the range of $[2,58]$ m. The Gaussian head is likewise implemented using a two-layer MLP and predicts the parameters of 48 Gaussian components for each instance. These components are converted to an occupancy field as

\begin{equation}
P_{q,k}(\mathbf{x})
=
1-\prod_{g=1}^{N_g}
\left[
1-\alpha_{q,g}
\exp\left(
-\frac{1}{2}
(\mathbf{x}-\mu_{q,g,k})^\top
\Sigma_{q,g}^{-1}
(\mathbf{x}-\mu_{q,g,k})
\right)
\right].
\end{equation}

Here, $\alpha_{q,g}\in[0,1]$ denotes the occupancy weight of each Gaussian.
The trajectory and confidence heads are also lightweight two-layer MLPs. The trajectory head predicts four displacement vectors corresponding to the future forecasting steps, while the confidence head estimates the existence probability of each query.

\subsection{Training Details}
\label{sec:appendix_training}

We train LighTROcc for 20 epochs using the AdamW optimizer~\citep{loshchilov2019iclr-decoupled} with an initial learning rate of $3\times10^{-4}$ and a weight decay of $0.01$. The learning rate is scheduled using cosine annealing. Training is conducted on eight NVIDIA RTX Pro 6000 Blackwell GPUs with a batch size of one per GPU.

For Hungarian matching, we use the 3D center distance, attention overlap cost, and confidence cost with weights of $10.0$, $0.3$, and $0.2$, respectively. 
For the training objective in Eq.~(8), we set $(\lambda_{\mathrm{conf}}, \lambda_{\mathrm{cen}}, \lambda_{\mathrm{depth}},
\lambda_{\mathrm{attn}}, \lambda_{\mathrm{occ}}, \lambda_{\mathrm{traj}})
= (1.0, 0.3, 1.0, 1.0, 1.0, 1.0)$.
For trajectory prediction, we augment each query feature with motion priors derived from the observed frames. Specifically, given three observed frames $(t-2,t-1,t)$, we compute two displacement vectors corresponding to $t-2 \rightarrow t-1$ and $t-1 \rightarrow t$. These observed displacements are concatenated with the query feature and fed to the trajectory head, which predicts four future displacement vectors. To improve training stability, we employ teacher forcing during early training and gradually transition to prediction-based training. Under teacher forcing, the motion priors are computed from ground-truth instance centers, whereas prediction-based training uses predicted centers. The teacher-forcing ratio is progressively reduced during warm-up and reaches zero from epoch 8 onward, after which all motion priors are derived from the predicted centers.

\section{Additional Experiments and Analysis}

\subsection{Flow-Based Instance Recovery in Dense Voxel-wise Models}
\label{sec:appendix_instance_recovery}
In this section, we further analyze the instance segmentation behavior of dense voxel-wise occupancy forecasting methods. As discussed in the main paper, methods such as Cam4DOcc~\citep{ma2024cam4docc}, EfficientOCF~\citep{xu2025efficientocf}, and OccProphet~\citep{chen2025occprophet} are not inherently designed in an instance-centric manner and therefore rely on an additional flow prediction module to recover object identities from dense occupancy maps. These methods typically predict backward centripetal flow, where each occupied voxel is encouraged to point toward the center of its corresponding object in the previous frame. The instance identity of a voxel is then determined according to the object center reached by its predicted flow. Voxels whose flows do not reach a valid object center cannot be reliably associated with any instance and may consequently remain unassigned. As a result, the instance segmentation quality of these dense methods is closely coupled with the accuracy of their flow predictions.

\begin{figure}[h]
    \centering
    \includegraphics[width=0.9\textwidth]{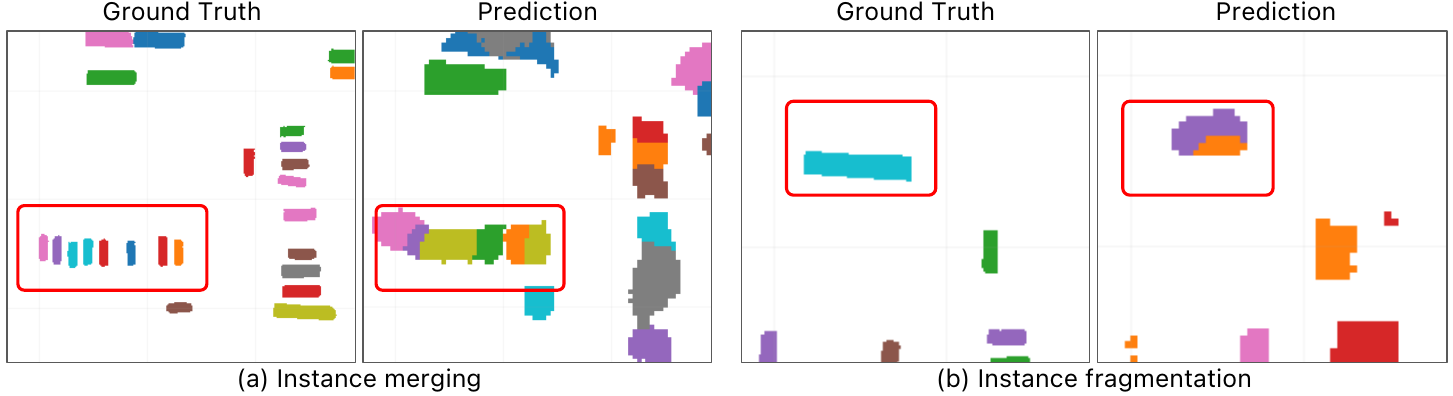}
    \vspace{-.7em}
    \caption{
        Qualitative examples of failure modes in flow-based instance recovery for dense voxel-wise models. (a) Instance merging, where voxels from multiple nearby objects are incorrectly assigned to a single identity. (b) Instance fragmentation, where voxels belonging to one object are split into multiple identities. Red boxes highlight representative errors.
    }
    \label{fig:appendix_flow_failure}
\end{figure}

As shown in Figure~\ref{fig:appendix_flow_failure}, even small flow errors can cause voxels to inherit incorrect instance identities, particularly in crowded scenes with nearby objects. This can merge multiple objects into a single instance or fragment one object into multiple identities, increasing both false-positive and false-negative predictions. Such association errors consequently degrade AP, which reflects both occupancy overlap and instance separation. These observations highlight the limitation of indirectly recovering object identities from dense occupancy and flow, and motivate the instance-centric design of LighTROcc, which maintains object-specific representations directly throughout forecasting.

\subsection{Temporal Shape Consistency}

A key advantage of LighTROcc is that its instance-centric representation preserves the predicted geometry of each object throughout the forecasting horizon. As discussed in Section~\ref{sec:appendix_instance_recovery}, dense voxel-wise methods recover instance identities indirectly from predicted flow. Flow errors can assign voxels to incorrect objects or leave them unassigned, resulting in fragmented instance masks and unstable object extents across frames. In contrast, LighTROcc propagates each predicted Gaussian mixture using the displacement vectors while retaining its component offsets, covariances, and occupancy weights, thereby preserving the shape and extent of each instance across future frames.

\begin{wraptable}[11]{r}{0.46\linewidth}
\vspace{-1.0em}
\centering
\caption{
Temporal shape consistency of 4D occupancy forecasting methods. Higher TSC indicates more stable geometry over time.
}
\label{tab:tsc}
\setlength{\tabcolsep}{6pt}
\begin{tabular}{lc}
\toprule
Method & TSC (\%) $\uparrow$ \\
\midrule
Cam4DOcc      & 51.04 \\
EfficientOCF  & \underline{65.67} \\
OccProphet    & 36.38 \\
\midrule
LighTROcc (Ours) & \textbf{91.32} \\
\bottomrule
\end{tabular}
\vspace{-8pt}
\end{wraptable}

Following UniOcc's object-aligned shape consistency evaluation~\citep{wang2025uniocc}, we compute Temporal Shape Consistency (TSC) to measure the preservation of predicted object geometry across consecutive frames, independent of global translation and orientation. Let $\hat{O}_{i}^{k}$ denote the occupied voxel set of instance $i$ at time step $k$, and $\mathcal{K}_i$ its evaluated consecutive time steps. After transforming each shape into an object-local coordinate system through centroid subtraction and PCA-based alignment, denoted by $\mathcal{A}(\cdot)$, the TSC is computed as
\begin{equation}
\mathrm{TSC}
=
\frac{1}{|\mathcal{I}|}
\sum_{i\in\mathcal{I}}
\frac{1}{|\mathcal{K}_i|}
\sum_{k\in\mathcal{K}_i}
\operatorname{IoU}\left(
\mathcal{A}(\hat{O}_{i}^{k}),
\mathcal{A}(\hat{O}_{i}^{k+1})
\right),
\end{equation}
where $\mathcal{I}$ denotes the set of evaluated predicted instances.

As shown in Table~\ref{tab:tsc}, dense voxel-wise baselines exhibit lower temporal shape consistency, reflecting greater variation in instance geometry across frames. LighTROcc achieves a TSC of 91.32, compared with 65.67 for EfficientOCF. This result reflects the temporal stability encouraged by the instance-centric representation.

\begin{table}[h]
\centering
\vspace{-1.0em}
\caption{
Forecasting performance across future horizons. LighTROcc (constant vel.) uses constant-velocity extrapolation instead of learned trajectory prediction. Best and second-best results are shown in bold and underlined, respectively.
}
\label{tab:forecasting_horizons}
\renewcommand{\arraystretch}{1.15}
\setlength{\tabcolsep}{4.5pt}

\resizebox{\linewidth}{!}{
\begin{tabular}{l c cccc cccc}
    \toprule

    \multirow{2}{*}{Method}
    & \multirow{2}{*}{Type}
    & \multicolumn{4}{c}{IoU}
    & \multicolumn{4}{c}{AP} \\

    \cmidrule(lr){3-6}
    \cmidrule(lr){7-10}

    &
    & 0.5 s
    & 1.0 s
    & 1.5 s
    & 2.0 s
    & 0.5 s
    & 1.0 s
    & 1.5 s
    & 2.0 s \\

    \midrule

    Cam4DOcc~\citep{ma2024cam4docc}
    & \multirow{3}{*}{\makecell{Dense\\Voxel-wise}}
    & 13.02
    & 12.11
    & 11.39
    & 10.68
    & 9.39
    & 8.17
    & 7.49
    & 7.03 \\

    EfficientOCF~\citep{xu2025efficientocf}
    &
    & 9.16
    & 8.77
    & 8.41
    & 8.02
    & 15.36
    & 14.62
    & 13.75
    & 12.78 \\

    OccProphet~\citep{chen2025occprophet}
    &
    & 16.21
    & \underline{14.46}
    & \underline{13.47}
    & \underline{12.78}
    & 10.67
    & 7.42
    & 5.95
    & 4.96 \\

    \midrule

    \textbf{LighTROcc (Ours)}
    & \multirow{2}{*}{Instance-wise}
    & \textbf{17.15}
    & \textbf{15.76}
    & \textbf{14.62}
    & \textbf{13.67}
    & \textbf{39.35}
    & \textbf{35.15}
    & \textbf{31.16}
    & \textbf{27.70} \\

    \textbf{LighTROcc (constant vel.)}
    &
    & \underline{16.30}
    & 12.72
    & 12.25
    & 11.18
    & \underline{35.13}
    & \underline{27.49}
    & \underline{23.66}
    & \underline{23.19} \\

    \bottomrule
\end{tabular}
}

\end{table}

\subsection{Performance at Different Forecasting Horizons}

To complement the aggregate results in Table~\ref{tab:main_results} of the main paper, we report IoU and AP at each future forecasting horizon. The current-only instance-wise baselines are excluded from this comparison because they do not explicitly forecast future occupancy states. As shown in Table~\ref{tab:forecasting_horizons}, all evaluated methods exhibit a gradual decrease in both IoU and AP as the forecasting horizon increases, reflecting the greater difficulty of longer-term occupancy prediction. Nevertheless, LighTROcc consistently achieves the best performance at every horizon, with particularly large margins in instance-level AP. Notably, at the longest horizon of $2.0\,s$, LighTROcc reaches an IoU of $13.67$ and an AP of $27.70$, outperforming the strongest prior baselines by 0.89 IoU and 14.92 AP points, respectively. These results demonstrate that the proposed instance-centric formulation retains a consistent forecasting advantage across all evaluated horizons.

To further isolate the contribution of the trajectory head, we replace the learned future displacements with constant-velocity extrapolation while retaining the same current instance predictions and Gaussian geometry. As shown in Table~\ref{tab:forecasting_horizons}, the constant-velocity variant consistently underperforms LighTROcc across all forecasting horizons. From 0.5 to 2.0~s, LighTROcc decreases from 17.15 to 13.67 IoU ($-$20.3\%) and from 39.35 to 27.70 AP ($-$29.6\%), whereas constant velocity decreases from 16.30 to 11.18 IoU ($-$31.4\%) and from 35.13 to 23.19 AP ($-$34.0\%). These results demonstrate the benefit of learned step-specific motion over linear extrapolation.

\begin{figure}[h]
    \centering
    \includegraphics[width=\textwidth]{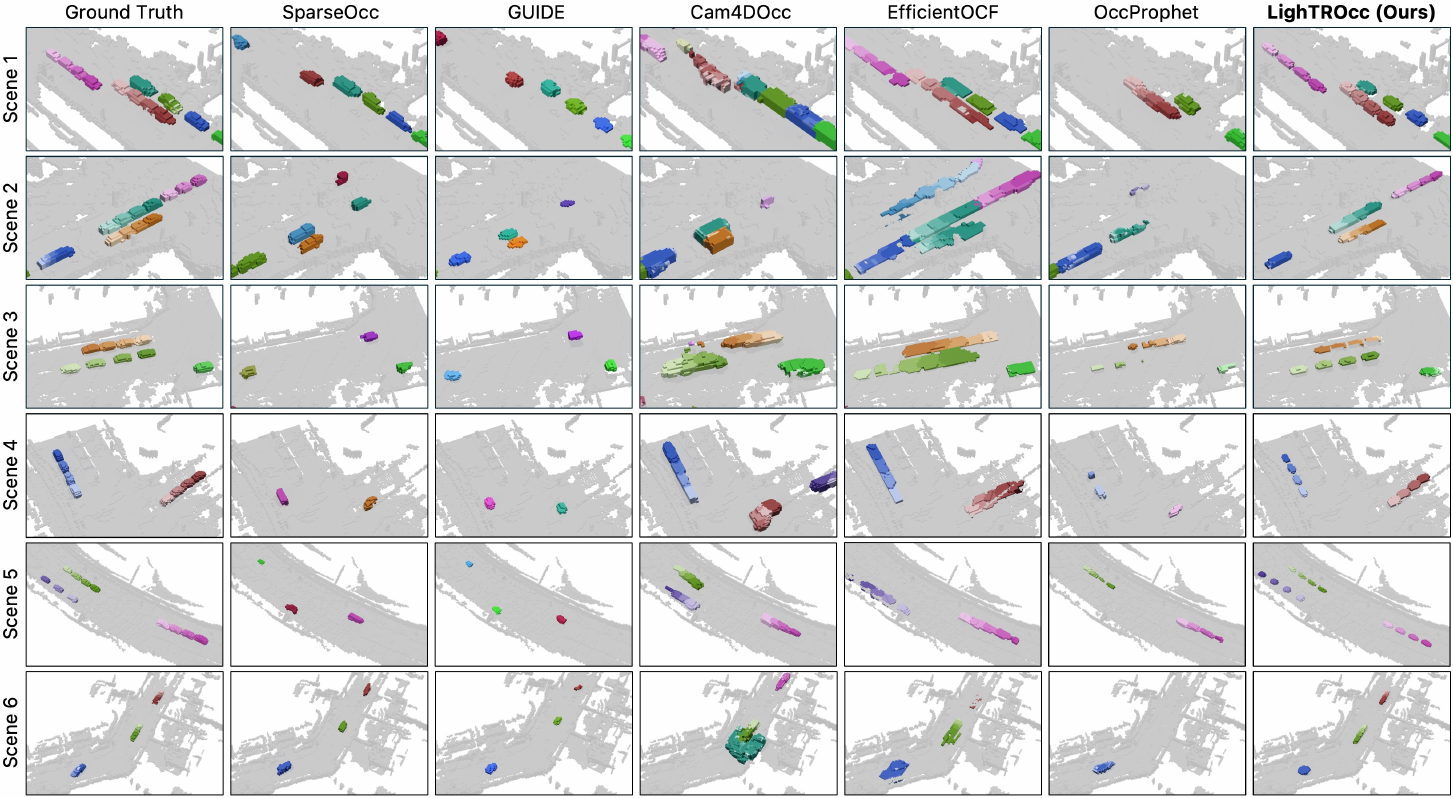}
    \vspace{-2.0em}
    \caption{
        Additional qualitative comparison of GMO occupancy predictions with a broader set of baselines. LighTROcc consistently captures object geometry and temporal evolution across diverse scenes. Colors become progressively darker for later forecasting steps.
    }
    \label{fig:appendix_qualitative}
\end{figure}

\section{Additional Qualitative Results}

Figure~\ref{fig:appendix_qualitative} provides additional qualitative comparisons with a broader set of baselines to complement the results in the main paper. For a comprehensive comparison, we visualize predictions from all evaluated baselines. Since SparseOcc~\citep{liu2024sparseocc} and GUIDE~\citep{hu2026guide} are current-only methods, their results are shown only for the present frame. Although these methods achieve competitive object localization at the current frame, they cannot model future occupancy evolution and therefore provide no predictions over the forecasting horizon. In contrast, the 4D forecasting methods explicitly reconstruct future occupancy states. Cam4DOcc~\citep{ma2024cam4docc} produces temporally extended predictions, visualized from lighter to darker colors toward later steps, but tends to overestimate object extents. EfficientOCF~\citep{xu2025efficientocf} exhibits a similar overestimation tendency and, as shown in Scene~2, can merge nearby objects into a single instance. OccProphet~\citep{chen2025occprophet} generally captures object extent more accurately, but its flow-based instance recovery is less stable, often causing object occupancy to progressively disappear at later forecasting steps. In comparison, LighTROcc consistently preserves instance geometry while accurately modeling future object evolution across diverse scenes. These qualitative results further demonstrate the effectiveness of the proposed instance-centric forecasting formulation.

\end{document}